\documentclass[11pt,a4paper]{article}
\usepackage{amsmath}
\usepackage{amsthm}
\usepackage{makeidx}
\usepackage{latexsym}
\usepackage{amssymb}

\usepackage[compatibility=false]{caption}
\usepackage{subcaption}

\usepackage{enumerate}

\usepackage{floatflt}
\usepackage{graphicx}
\graphicspath{{./img/}}
\DeclareGraphicsExtensions{.pdf,.jpg,.jpeg,.png}

\usepackage{bm}
\usepackage{epsfig}
\usepackage{epsf}

\usepackage{hyperref}
\usepackage{float}
\usepackage{geometry}
\theoremstyle{remark}

\title{A fuzzy approach for segmentation of touching characters}
\author{Giuseppe Air\`{o} Farulla$^1$, Nadir Murru$^2$, Rosaria Rossini$^3$\\ \\
\small{giuseppe.airofarulla@polito.it, nadir.murru@unito.it, rossini@ismb.it}\\ \\
\small{$^1$Department of Control and Computer Engineering, Politecnico di Torino,}\\
\small{Corso Duca degli Abruzzi 24, 10129, Torino, Italy}\\
\small{$^2$Department of Mathematics, University of Turin,}\\
\small{Via Carlo Alberto 10, 10121 Torino, Italy}\\
\small{$^3$Istituto Superiore Mario Boella, Center for Applied Research on ICT,}\\
\small{Via Pier Carlo Boggio 61, 10138, Torino, Italy}
}
\date{}

\begin{document}

\maketitle

\begin{abstract}
The problem of correctly segmenting touching characters is an hard task to solve and it is of major relevance in pattern recognition. In the recent years, many methods and algorithms have been proposed; still, a definitive solution is far from being found. In this paper, we propose a novel method based on fuzzy logic. The proposed method combines in a novel way three features for segmenting touching characters that have been already proposed in other studies but have been exploited only singularly so far. The proposed strategy is based on a 3--input/1--output fuzzy inference system with fuzzy rules specifically optimized for segmenting touching characters in the case of Latin printed and handwritten characters. The system performances are illustrated and supported by numerical examples showing
that our approach can achieve a reasonable good overall accuracy in segmenting characters even on tricky conditions of touching characters.
Moreover, numerical results suggest that the method can be applied to many different datasets of characters by means of a convenient tuning of the fuzzy sets and rules.
\end{abstract}

\section{Introduction}
Automatic recognition of both printed and handwritten characters remains a challenging problem in pattern recognition.
Most existing Optical Character Recognition software (OCR) deal with it by exploiting simultaneously two highly correlated techniques: character segmentation and pattern recognition.
As part of the OCR process, character segmentation techniques are applied to patterns representing individual characters to be recognized. The simplest way to perform character segmentation would be to exploit the space between characters.
This strategy unfortunately fails when considering mathematical formulae, handwritten and printed words with touching characters, representing well known issues that often occur in degraded (e.g., photocopies) or compressed text images \cite{roy2012multi}.
In these situations, two or more adjacent characters touch together and share common pixels.
To identify the touching regions and provide a correct segmentation is crucial to recognition, since incorrectly segmented characters are unlikely to be correctly recognized even from high-performance pattern recognition algorithms \cite{zhao2003two}.
In facts, many researchers state that errors in characters segmentation affect overall pattern recognition performance more than a degradation in the starting image \cite{Jung}. 
Segmentation of touching components is crucial to get higher recognition rates by OCR systems \cite{Casey}.

%

Common techniques for character segmentation exploit several aspects characterizing letters and their shapes, such as vertical projection, pitch estimation or character size, contour analysis, or segmentation--recognition coupled techniques \cite{Lu}, \cite{Lu2}.
One of the most difficult problem an image segmentation algorithm has to address is the segmentation of touching characters \cite{Roy}, \cite{Saba}.
Very often, adjacent characters are touching, and may overlap, making hard the task of segmenting a given expression or word
correctly into its character components \cite{bansal2002segmentation,liang1994segmentation}.
Given the relevance of such challenging task, several methods have been developed in last years for performing optimal segmentation of touching characters. Kurniawan et al. \cite{Kur} identify touching positions in Latin handwritten characters by means of self organizing feature maps and a region--based approach. In \cite{LeeVerma} and \cite{Liang}, the authors also deal with segmentation of Latin handwritten texts.
Different approaches involving thinning algorithms can be found in \cite{Lu3} and \cite{Pra}.
Among the earlier pieces of work on touching character segmentation, some approaches rely on contour analysis of the
connected components for segmentation \cite{kim2000recognition,Fle}.
In \cite{Sha}, Sharma et al. study the problem of detecting arbitrarily--oriented text from video frames. Cut positions in touching characters are evaluated using the top distance profile.
Roy et al. \cite{Roy2} addressed the problem of segmenting touching characters with different orientations.
In \cite{Wei} and \cite{Saba2}, authors developed segmenting approaches leveraging on genetic algorithms. 
Rehman et al. \cite{Reh} identify character boundaries by using a set of heuristic rules. 
Louloudis et al. \cite{Lou} performed text line and word segmentation of handwritten documents by applying the Hough transform.
In \cite{Manma}, the authors addressed the problem of automatically segmenting words from historical handwritten documents.
The water reservoir algorithm has been exploited in some researches, e.g., \cite{Pal} and \cite{Kumar}. 
In \cite{Ban} and \cite{Sta}, the authors presented methods derived by combining different segmentation techniques.
Further methods can be found in \cite{Alex}, \cite{Cama}, \cite{Frink}, \cite{He} and \cite{Weinman}. 
It is worth mentioning that works \cite{Infty} and \cite{Utpal} also deal with segmenting characters and symbols within mathematical expressions.

Generally, these strategies need a preprocessing step where the input RGB image is converted to grayscale by eliminating the hue and saturation values while retaining the illumination, then converting the grayscale image to binary, obtaining a matrix whose entries are 0 for foreground pixels (black) and 1 for background pixels (white) for all other pixels. Since different thresholds are used for detection it is possible that some feature of characters is lost in the process. In our experiments we resort on the Otsu thresholding method \cite{Otsu}, which has proven to be very robust to noise and to changes in scenes and input images, providing thresholds for image binarization ensuring that information loss is minimized.

A function is used to evaluate each column of the matrix leveraging on features that characterize typical character positions within the text, giving a value to each column. Finally, cut positions (i.e., columns) are chosen depending on these values.
Common functions implied are the ratio of the second difference of the vertical projection (e.g., \cite{Kahan}) and the peak--to--valley (e.g., \cite{Lu4}). Other functions are, e.g., based on number of black pixels, number of white pixels counted from the top of the column to the first black pixel, crossing count (i.e., number of black to white transitions),  number of identical black (white) pixels with left (right) column, width to height ratio for the remaining left (right) pattern after cutting (e.g., \cite{Bayer}). Another approach can be found in \cite{Utpal2}, where the authors used the inverse crossing count, measure of blob thickness and degree of ''middleness'' for defining a function that identifies when a column, a row, or a diagonal is a cutting position.

However, we argue that all the methods and approaches above mentioned do not provide a comprehensive answer to the problem of segmenting touching characters. Indeed, their performances are not always optimal, radically depending on the specific set of characters involved in the segmentation. At the moment, there does not exist a standard approach for the segmentation of touching characters. Thus, this is currently an active research field.

There has always been a dilemma whether it is more convenient to segment first and then recognize the patterns, or instead classify while segmenting. Authors in \cite{bansal2002segmentation} review Casey and Lecolinet work \cite{Casey} stating that that the strategies for segmentation can be classified into three main strategies as follows:
\begin{enumerate}
\item the classical approach already described;
\item recognition-based segmentation, in which a search is made for image components that match with the character classes in a valid alphabet;
\item holistic approach that attempts to recognize the word as a whole.
\end{enumerate}
In this classification, our work lies near to the first class, still presenting some important advances: in fact, we aim at combining some of the previously cited features, usually exploited one at a time, by means of an original fuzzy logic approach in order to improve performances in separating touching characters. Indeed, the selection of the features that characterize touching positions is an art rather than a technique. In other words, the selection of the features mainly depends on the experience of the authors. Thus, in this context, fuzzy logic can be very useful, since it is congenial to capture and to code expert--based knowledge in view of performing targeted simulations. Taking this into strong consideration, we also leverage on optimization techniques to increase overall performances of our approach.

Fuzzy logic has been already exploited to perform image segmentation. For instance, Garain and Chaudhuri \cite{Utpal2} used fuzzy multifactorial analysis to combine some of the features previously described. In \cite{Naz}, a survey on image segmentation techniques using fuzzy clustering is presented. Fuzzy logic has been also exploited for developing segmentation--recognition coupled techniques \cite{Heb}. A non--linear fuzzy approach can be found in \cite{Sarkar}. 
In \cite{Nac}, authors used edge corners and fuzzy logic to develop segmentation techniques exploited to break down Captcha. Further approaches can be found in \cite{Jay}, \cite{Shi}, \cite{Tobias}.

In this paper, we propose a novel fuzzy approach that differs from the state of the art in several aspects. Firstly, we combine by means of a fuzzy strategy some features that have never been exploited together in previous works. Secondly, we develop an original strategy based on an inference system composed by 3--input/1--output with fuzzy rules specifically optimized for the purpose of separating touching characters in the case of Latin printed and handwritten characters. The strength of our fuzzy strategy relies on the possibility to adjust its parameters in such a way that they can fit the characteristics of the data set. In other words, the parameters of the method are extracted a priori considering the different characters in the data set.

The inference engine is based on the Mamdani model with if--then rules, minimax set--operations, sum for composition of activated rules and defuzzification based on the centroid method. We have chosen the Mamdani model since it is congenial to capture and to code expert--based knowledge \cite{Mam}. 

The paper is organized as follows. In Section \ref{sec:fuzzy}, we present the fuzzy strategy conveniently developed for performing segmentation of touching characters. In section \ref{sec:test}, we present the numerical results that show the effectiveness of the proposed method. Specifically, in Section \ref{sec:dataset}, we describe the datasets used for simulations. Sections \ref{sec:test-print} and \ref{sec:test-hand} are devoted to test the method on datasets of Latin printed and handwritten characters, respectively. Finally, in Section \ref{sec:conc} we draw some conclusion and present future works.

\section{Fuzzy strategy}\label{sec:fuzzy}

In the following, we only focus on binarized images, for homogeneity with the solutions already presented. In a binarized image, a pattern can be represented by a matrix whose entries are 0 (black pixels) and 1 (withe pixels). Generally, methods for segmenting touching characters define a function based on some features that characterize cut positions. Then, such a function is evaluated for each column of the matrix and the cut position is chosen depending on these values. Classical functions of this kind are the peak--to--valley function $g$ and the function $h$ defined as
$$g(i)=\cfrac{V(l_i)-2V(i)+V(r_i)}{V(i)+1},\quad h(i)=\cfrac{V(i-1)-2V(i)+V(i+1)}{V(i)},$$
where $V(i)$ denotes the vertical projection function for the $i$-th column, $l_i$ and $r_i$ are the peak positions on the left side and right side of $i$, respectively. The column with the highest value of $g$ (or $h$) is identified as the cutting column. A further feature, that can suggests if the $i$--th column can be a cut position, is the distance $f(i)$ between $i$ and the center of the pattern. Indeed, generally, cutting columns are located near to the center of the pattern. Clearly, this feature should be only considered as an indication of the neighborhood where the cutting column is probably located. Indeed, we will exploit such a feature in combination with the previous functions with the aim of use it in order to correct the results provided by the other functions. In this section, we combine functions $f$, $g$, and $h$ by means of a fuzzy strategy that conveniently balances these functions.
 
Let us introduce the notion of a ``fuzzy degree" qualifying a column $i$ to be a cut position: in short, $\rho = \rho(i) \in [0,1]$. In our model, the lower the value of $\rho$, the more probable is that we have located a good cutting position. The strategy can be detailed by means of the fuzzification of the functions $f$, $g$, $h$.

Given a pattern in a binarized image, let $A$, $m$, $n$, and $c$ be the matrix of pixels of the binarized image, the number of row of $A$, the number of column of $A$, and the central column of $A$, respectively. In the following, when we refer to a column $i$ of $A$, we refer to the $i$--th column of $A$, i.e., we are considering the vector of length $m$ whose elements are the entries of the $i$--th column or we are only considering its position. This will be clear from the context.

The central column $c$ is evaluated by means of $c=\frac{n+1}{2}$. When $n$ is odd, $c$ is clearly the central column of $A$; when $n$ is even, we consider as the central column the mean between $\frac{n}{2}$--th column and $\frac{n}{2}+1$, even if in this case $c$ is not an integer number. In this way, for each column $i$ of $A$, we define its distance from the center of the pattern as $f(i)=\lvert c-i \rvert$. In our fuzzy strategy, we take into account the normalized distance between each column $i$ and the central column $c$, i.e., we consider $\bar f(i)=\frac{f(i)}{c}$. 

Similarly, for each column $i$ of $A$, instead of directly using the functions $g$ and $h$, we consider the normalized functions
$$\tilde g(i)=\cfrac{g(i)-\min_{j\in\mathcal C} g(j)}{\max_{j\in\mathcal C} g(j)-\min_{j\in\mathcal C} g(j)},\quad \tilde h(i)=\cfrac{h(i)-\min_{j\in\mathcal C} h(j)}{\max_{j\in\mathcal C} h(j)-\min_{j\in\mathcal C} h(j)},$$
where $\mathcal C=\{1,2,...,n\}$ is the set of the columns of $A$. Note that functions $\tilde g$ and $\tilde h$ are well--defined since we consider matrices $A$ where at least two columns are different. Finally, in the following we will use the functions
$$\bar g = 1 - \tilde g, \quad \bar h = 1 - \tilde h$$
so that low values of $\bar g$ and $\bar h$ identify cutting columns.

Functions $\bar f$, $\bar g$, $\bar h$ are fuzzified by defining convenient fuzzy sets and related membership functions. The fuzzy degree $\rho$ will be evaluated combining these functions by means of some fuzzy rules. Fuzzy sets, membership functions, fuzzy rules will be specified in sections \ref{sec:test-print} and \ref{sec:test-hand}.

The inference engine will be the basic Mamdani model \cite{Mam}, with if--then rules, minimax set--operations, sum for composition of activated rules, and defuzzification based on the centroid method.  The Mamdani model is congenial to capture and to code expert--based knowledge in view of performing targeted simulations; accordingly the system's performance is tuned by means of expert--based choices, heuristic criteria and non--linear optimization methods.

\section{Experimental tests}\label{sec:test}

Although creating a specific dataset of touching character is not our main contribution, in our research we faced the lacking of good and standardized datasets on which testing our algorithms. This reason forced us to build a dataset of touching characters.
This need is shared with many other researchers and works, including recent ones as \cite{sadri2016novel} (although this last paper refers to Persian handwriting recognition algorithms rather than to Latin characters-based ones).
In particular, we created two different datasets for Latin characters, one containing handwritten cursive characters, the other containing sided machine printed characters.
In the following, we discuss our choices and methodology. 

Due to the different characteristics of the two datasets, in the remaining parts of the section we discuss the process to select the best  parameters for running our method on the two datasets. Our fuzzy strategy has specific parameters meant to capture the differences between one dataset to another. Such parameters are tailored on the specific dataset characteristic by using an heuristic way and non--linear optimization methods. Moreover, we would like to point out that the three features combined in our fuzzy routine seem to be sufficient for obtaining optimal results in the segmentation. In facts, adding further features appear to be useless. For instance, we have verified that the use of the crossing count as a fourth fuzzy input did not lead to improvements. The experimental results are presented in the following.

\subsection{Construction of the datasets}\label{sec:dataset}
The ultimate purpose of getting good segmentation results, especially when touching character are considered, is to boost the the performances for what concerns overall recognition accuracy. Despite the fact thus that, to be fair, different approaches should be evaluated on the basis of their recognition performances, there are not many authors publishing their results on a benchmark database \cite{lee2012binary}.
Also, as stated in many state-of-the-art work, e.g., \cite{Kur}, unfortunately at the present a comprehensive dataset specific on touching characters is still missing. Given that, for researches it is difficult to conduct experiments and to analyze any proposed method; in addition, it is hard to fair compare performances and results obtained from different authors.
We try to overtake such an obstacle by proposing two datasets.

\begin{figure*}[t!]
    \centering
    \begin{subfigure}[t]{0.2\textwidth}
        \centering
        \includegraphics[width=\textwidth]{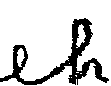}
        \caption{\label{fig:eh}}
    \end{subfigure}%
    ~ 
    \begin{subfigure}[t]{0.29\textwidth}
        \centering
        \includegraphics[width=\textwidth]{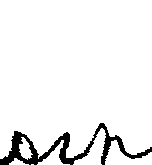}
        \caption{\label{fig:arn}}
    \end{subfigure}%
    ~ 
    \begin{subfigure}[t]{0.245\textwidth}
        \centering
        \includegraphics[width=\textwidth]{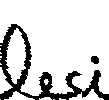}
        \caption{\label{fig:lesi}}
    \end{subfigure}
    \caption{\label{fig:samples-hand}Samples of, respectively, two (\subref{fig:eh}), three (\subref{fig:arn}), and four (\subref{fig:lesi}) handwritten cursive touching character patterns.}
\end{figure*}

The first one, called dataset A, contains images of handwritten cursive characters we have built relying on samples from a standard dataset, in fashion of \cite{Kur}. In particular we started from the CCC database \cite{Cama}. The CCC database contains 57'293 samples of cursive characters that were manually extracted from images coming from different input sources, mainly related to American Post Services. They include both upper and lower case letters. Each sample is stored as a binary matrix within the database, and accompanied with information about the size of the matrix itself and the character that is represented.
Starting from the whole database, we developed a MATLAB script to randomly extract 1'000 of its samples, taking care of maintaining a uniform distribution for all the characters chosen, both in their upper and lower version.
These samples were later combined and merged together to form two, three, and four touching character patterns, each of whom is accompanied by a textual descriptor indicating the index of the proper cut column (or columns, in the case of three and four multiple touching character patterns).
One sample of each category of patterns is shown in \figurename~\ref{fig:samples-hand}. For instance, the descriptor of the sample represented in \figurename~\ref{fig:eh} states that the proper column to cut to properly separate the ``e" and the ``h" characters is the $52^{th}$.

This merging is however unsupervised, and so improper combinations happen during the process. So, we had to filter out the most unrealistic ones. Firstly, we discarded all the samples with significant difference in their heights. Secondly, we manually removed combinations without touching patterns (i.e., the characters were well separated) or with touching patterns that seemed impossible to happen in real world.
At the end, we kept 153 combinations, of which 139 represent two touching character patterns, and the other are equally divided into three and four touching character patterns. The disproportion because the common touching characters consist of two characters, while three or more touching characters are rare \cite{Wei}. Moreover, note that the quantity of combinations of touching characters contained in our dataset is in compliance with other similar datasets constructed using the CCC database (e.g., in \cite{Kur} a dataset of 123 touching characters is used).

\begin{figure*}[t!]
    \centering
    \begin{subfigure}[t]{0.2\textwidth}
        \centering
        \includegraphics[width=\textwidth]{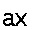}
        \caption{\label{fig:ax}}
    \end{subfigure}%
    ~ 
    \begin{subfigure}[t]{0.243\textwidth}
        \centering
        \includegraphics[width=\textwidth]{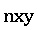}
        \caption{\label{fig:nxy}}
    \end{subfigure}%
    ~ 
    \begin{subfigure}[t]{0.31\textwidth}
        \centering
        \includegraphics[width=\textwidth]{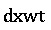}
        \caption{\label{fig:dxwt}}
    \end{subfigure}
    \caption{\label{fig:samples-machine}Samples of, respectively, two (\subref{fig:eh}), three (\subref{fig:nxy}), and four (\subref{fig:dxwt}) machine printed touching character patterns.}
\end{figure*}

The second one, called dataset B, contains images of sided machine printed characters we have built resorting on a second MATLAB script. 
We have identified a list of font types (namely Cambria, Candara, Georgia, Lucida Sans Regular, Times New Roman and Verdana Bold) and sizes (namely 10, 20 and 25); for each type and size a MATLAB script combines into images the lower characters from the alphabet to form two, three, and four touching character patterns. Each image is accompanied by a textual descriptor, which in this case indicates directly the characters represented.
One sample of each category of patterns is shown in \figurename~\ref{fig:samples-machine}. For instance, the descriptor of the sample represented in \figurename~\ref{fig:ax} states that the images represent the string ``ax".
Also in this case, we preferred to revise manually the dataset to remove missing, or unrealistic, touching patterns. At the end, we kept the most promising 189 combinations (where 168 are composed by two touching characters), in order to define a challenging dataset to test our approach.

\subsection{Tests on Latin printed characters}\label{sec:test-print}
In the following, we discuss the results of segmentation of touching characters from the dataset B described in the previous section, accordingly to the fuzzy strategy described in section \ref{sec:fuzzy}. Fuzzy sets and membership functions related to $\bar f$, $\bar g$ and $\bar h$ are defined accordingly to expert based choices. Moreover, their construction has been optimized by using the Particle Swarm Optimization (PSO) algorithm \cite{PSO} in order to improve overall performances of our fuzzy strategy. Similarly, the fuzzy rules (described in the following) have been tuned using both heuristic criteria and the PSO algorithm.

Given a matrix $A$ as defined in section \ref{sec:fuzzy}, for each column $i$ of $A$, $\bar f(i)$, $\bar g(i)$ and $\bar h(i)$ are evaluated and the degree $\rho(i)$ is provided by the following inference scheme that includes three inputs (fuzzification of $\bar f$, $\bar g$, $\bar h$) and one output (cutting degree $\rho$). The column $i$ with the lowest value of $\rho$ is considered as the cut column.

The function $\bar f$ is fuzzified by defining the following fuzzy sets:
\begin{itemize}
\item if $\bar f(i)\leq 0.35$, then distance from the center of the pattern is \emph{Low};
\item if $\bar 0.15\leq f(i)\leq 0.75$, then distance from the center of the pattern is \emph{Medium};
\item if $\bar f(i)\geq 0.5$, then distance from the center of the pattern is \emph{High}.
\end{itemize}

For the function $\bar g$, we define the following fuzzy sets:
\begin{itemize}
\item if $\bar g(i)\leq 0.4$, then $\bar g(i)$ is \emph{Low};
\item if $0.2\leq \bar g(i)\leq 0.5$, then $\bar g(i)$ is \emph{Medium};
\item if $\bar g(i)\geq 0.45$, then $\bar g(i)$ is \emph{High}.
\end{itemize}

The function $\bar h$ is fuzzified by means of the following fuzzy sets:
\begin{itemize}
\item if $\bar h(i)\leq 0.4$, then $\bar h(i)$ is \emph{Low};
\item if $\bar 0.1\leq h(i)\leq 0.75$, then $\bar h(i)$ is \emph{Medium};
\item if $\bar h(i)\geq 0.5$, then $\bar h(i)$ is \emph{High};
\end{itemize}

\begin{figure}[p] 
\centering
\includegraphics[scale=0.9]{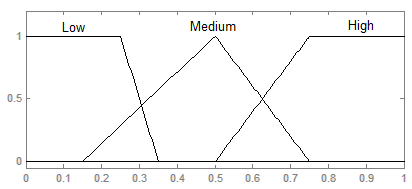}
\caption{Membership functions of the fuzzy sets related to $\bar f$ (for dataset B)}
\label{fig:mf-f}
\end{figure}

\begin{figure}[tp] 
\centering
\includegraphics[scale=0.9]{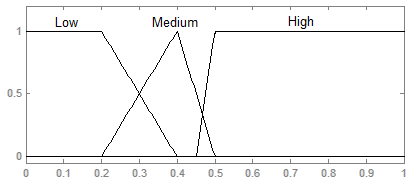}
\caption{Membership functions of the fuzzy sets related to $\bar g$ (for dataset B)}
\label{fig:mf-g}
\end{figure}

\begin{figure}[tp] 
\centering
\includegraphics[scale=0.9]{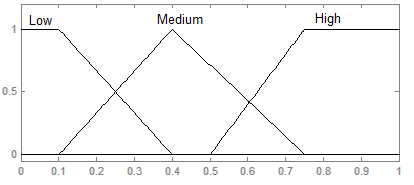}
\caption{Membership functions of the fuzzy sets related to $\bar h$ (for dataset B)}
\label{fig:mf-h}
\end{figure}

Figures \ref{fig:mf-f}, \ref{fig:mf-g}, and \ref{fig:mf-h} show the membership functions of the previous fuzzy sets.

\begin{figure}[tp] 
\centering
\includegraphics[scale=0.9]{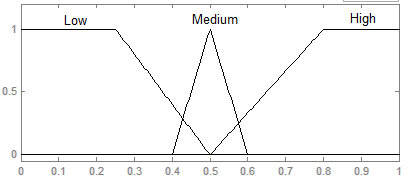}
\caption{Membership functions of the fuzzy sets related to $\rho$ (for dataset B)}
\label{fig:mf-o}
\end{figure}

Finally, for the fuzzy output $\rho$ we define the following fuzzy sets, whose membership functions are depicted in Figure \ref{fig:mf-o}:
\begin{itemize}
\item if $\rho(i)\leq 0.5$, then $\rho(i)$ is \emph{Low};
\item if $0.4\leq \rho(i)\leq 0.6$, then $\rho(i)$ is \emph{Medium};
\item if $\rho(i)\geq 0.5$, then $\rho(i)$ is \emph{High};
\end{itemize}

The inference system is based on the following rules, that combine the three inputs $\bar f(i), \bar g(i), \bar h(i)$ in order to produce the fuzzy output $\rho(i)$, for each column $i$ of $A$:
\begin{enumerate}
\item if $\bar f(i)$ is Low and $\bar h(i)$ is Low, then $\rho(i)$ is Low;
\item if $\bar f(i)$ is Low and $\bar g(i)$ is not High and $\bar h(i)$ is not Low, then $\rho(i)$ is Low;
\item if $\bar f(i)$ is Low and $\bar g(i)$ is High and $\bar h(i)$ is Medium, then $\rho(i)$ is Medium;
\item if $\bar f(i)$ is Medium and $\bar h(i)$ is not High, then $\rho(i)$ is Medium;
\item if $\bar f(i)$ is Medium and $\bar g(i)$ is Low and $\bar h(i)$ is High, then $\rho(i)$ is Medium;
\item if $\bar f(i)$ is High and $\bar g(i)$ is not High and $\bar h(i)$ is Low, then $\rho(i)$ is Medium;
\item if $\bar f(i)$ is High and $\bar g(i)$ is Low and $\bar h(i)$ is Medium, then $\rho(i)$ is Medium;
\item if $\bar f(i)$ is Low and $\bar g(i)$ is High and $\bar h(i)$ is High, then $\rho(i)$ is High;
\item if $\bar f(i)$ and $\bar g(i)$ and $\bar h(i)$ are not Low, then $\rho(i)$ is High;
\item if $\bar f(i)$ and $\bar g(i)$ are High, then $\rho(i)$ is High.
\end{enumerate}

The touching characters in dataset B are correctly segmented in the $96.1\%$ of the cases. For evaluating the correctness of the segmentation, we used a pattern recognition algorithm constructed by a neural network trained on the characters that compose the dataset B, in fashion of what done by other works reviewed in \cite{zhou2002verification}, and obtaining comparable results. We consider that touching character are correctly segmented when the pattern recognition algorithm correctly recognizes the characters after the segmentation.

Simulations show that the fuzzy combination of the functions $f, g, h$ improves the correct identification of the cutting column with respect to their separated use.

\begin{figure}[tp] 
\centering
\includegraphics[scale=0.4]{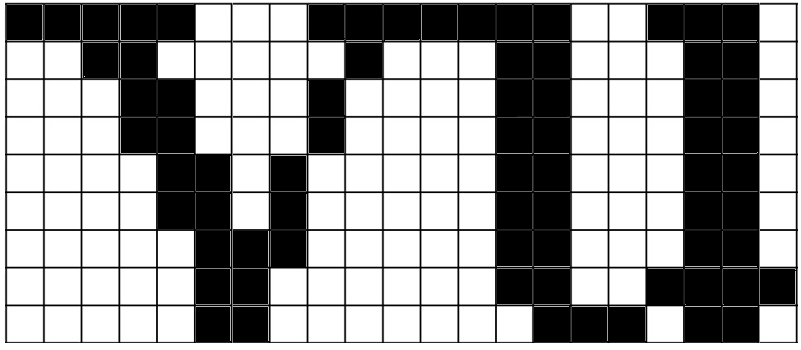}
\caption{Touching characters ``vu'' for font Times New Roman and font size of 20}
\label{fig:vu}
\end{figure}

\begin{figure}[tp] 
\centering
\includegraphics[scale=0.5]{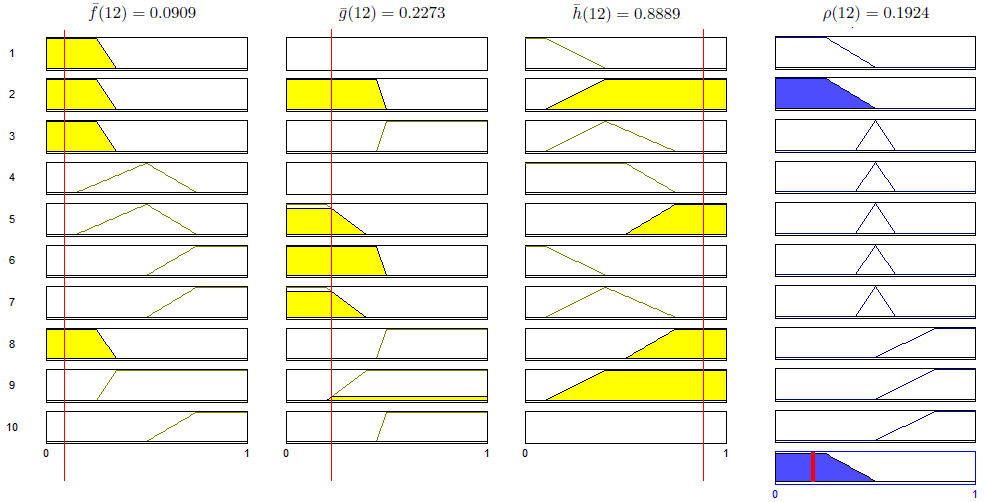}
\caption{Application of the fuzzy inference system to the column 12 of the pattern ``vu''}
\label{fig:vu12}
\end{figure}

To assess the performances of the fuzzy strategy compared to the usage on only the functions $g$ and $h$, a numerical example is reported below. Let us consider the touching characters ``vu'' (font Times New Roman) depicted in Figure \ref{fig:vu}. Our fuzzy routine correctly identifies the cutting column as the column 12 which is assigned the minimum value of $\rho$ among all the columns of the pattern. Specifically, we obtain $\rho(12) = 0.1924$. The fuzzy procedure performance is shown in Figure \ref{fig:vu12}, with application to the column 12. On the other hand, both $g$ and $h$ separately locate the column 16 as the cutting column. Indeed, we can observe that, e.g., $h(16) = 8$ that is greater than $h(i)$, for $i=1,...,21$, $i\not=16$, for example $h(12) = 0$. In Table \ref{table:vu}, we report the values of $\bar f, \bar g, \bar h, \rho$ for each column of the previous pattern (except for the first and the last column that are not surely cutting columns). 

\begin{table}[tp]\small
\caption{Values of $\bar f$, $\bar g$, $\bar h$, $\rho$ for the columns of the touching characters ``vu'' (excluded first and last column)}
\centering
\tabcolsep=0.15cm
\scalebox{0.82}{
\begin{tabular}{|c||c|c|c|c|}
\hline 
Column  & $\bar f$ & $\bar g$ & $\bar h$ & $\rho$  \cr \hline \hline
2 & 0.8182 & 0.4545 & 0.7778 & 0.7849 \cr \hline 
3 & 0.7273 & 0.6818 & 0.8333 & 0.8123 \cr \hline
4 & 0.6364 & 0.8409 & 0.9167 & 0.7908 \cr \hline
5 & 0.5455 & 0.8523 & 0.9111 & 0.8073 \cr \hline
6 & 0.4545 & 0.8333 & 0.9333 & 0.8102 \cr \hline
7 & 0.3636 & 0.6818 & 0.8148 & 0.7949 \cr \hline
8 & 0.2727 & 0.6818 & 0.8889 & 0.8047 \cr \hline
9 & 0.1818 & 0.6818 & 0.9259 & 0.8169 \cr \hline
10 & 0.0909 & 0.5303 & 0.8889 & 0.8169 \cr \hline
11 & 0 & 0.2273 & 0.7778 & 0.1984 \cr \hline
12 & 0.0909 & 0.2273 & 0.8889 & 0.1924 \cr \hline
13 & 0.1818 & 0.2273 & 0.1111 & 0.2078 \cr \hline
14 & 0.2727 & 0.9343 & 0.9722 & 0.8047 \cr \hline
15 & 0.3636 & 0.9205 & 1 & 0.7949 \cr \hline
16 & 0.4545 & 0 & 0 & 0.5000 \cr \hline
17 & 0.5455 & 0 & 0.7778 & 0.6305 \cr \hline
18 & 0.6364 & 0.3788 & 0.9556 & 0.7302 \cr \hline
19 & 0.7273 & 0.9091 & 0.9753 & 0.8123 \cr \hline
20 & 0.8182 & 1 & 0.9877 & 0.8169 \cr \hline
 
\end{tabular}
}
\label{table:vu}
\end{table}

\subsection{Tests on Latin handwritten characters}\label{sec:test-hand}
In the following, we perform segmentation of touching characters from the dataset. Similarly to what described in the previous section, fuzzy sets, membership functions, and fuzzy rules have been defined accordingly to expert based choices and further optimized leveraging the PSO algorithm. Patterns in the dataset A are greatly different from the ones in dataset B. For instance, touching positions in dataset B are often near to the center of the pattern. In the case of dataset A, cutting columns may occur more frequently at high distance from the center. On the other hand, the peak to valley function seems to have better performances in the case of the dataset A. Taking this into account, the optimization conducted by using the PSO algorithm has been strategic as it allowed us to highlight properties and connections among functions $f, g, h$ which are not noticeable at a glance. All these features are reflected in the following definition of fuzzy sets, membership functions, and fuzzy rules.

The fuzzy sets related to $\bar f$ are defined by
\begin{itemize}
\item if $\bar f(i)\leq 0.45$, then distance from the center of the pattern is \emph{Low};
\item if $\bar 0.25\leq f(i)\leq 0.55$, then distance from the center of the pattern is \emph{Medium};
\item if $\bar f(i)\geq 0.5$, then distance from the center of the pattern is \emph{High}.
\end{itemize}

For the function $\bar g$, we define the following fuzzy sets:
\begin{itemize}
\item if $\bar g(i)\leq 0.2$, then $\bar g(i)$ is \emph{Low};
\item if $0.15\leq \bar g(i)\leq 0.55$, then $\bar g(i)$ is \emph{Medium};
\item if $\bar g(i)\geq 0.25$, then $\bar g(i)$ is \emph{High}.
\end{itemize}

The fuzzy sets related to $\bar h$ are defined by
\begin{itemize}
\item if $\bar h(i)\leq 0.3$, then $\bar h(i)$ is \emph{Low};
\item if $\bar 0.15\leq h(i)\leq 0.65$, then $\bar h(i)$ is \emph{Medium};
\item if $\bar h(i)\geq 0.5$, then $\bar h(i)$ is \emph{High};
\end{itemize}

\begin{figure}[tp] 
\centering
\includegraphics[scale=0.9]{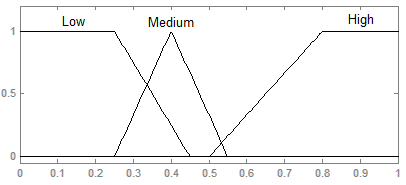}
\caption{Membership functions of the fuzzy sets related to $\bar f$ (for dataset A)}
\label{fig:mf-f-ccc}
\end{figure}

\begin{figure}[tp] 
\centering
\includegraphics[scale=0.9]{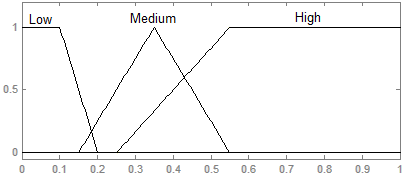}
\caption{Membership functions of the fuzzy sets related to $\bar g$ (for dataset A)}
\label{fig:mf-g-ccc}
\end{figure}

\begin{figure}[tp] 
\centering
\includegraphics[scale=0.9]{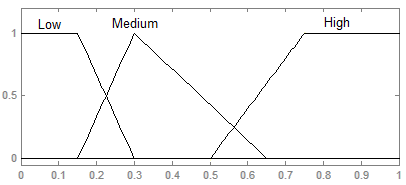}
\caption{Membership functions of the fuzzy sets related to $\bar h$ (for dataset A)}
\label{fig:mf-h-ccc}
\end{figure}

Figures \ref{fig:mf-f-ccc}, \ref{fig:mf-g-ccc}, and \ref{fig:mf-h-ccc} show the membership functions of the previous fuzzy sets. 

\begin{figure}[tp] 
\centering
\includegraphics[scale=0.9]{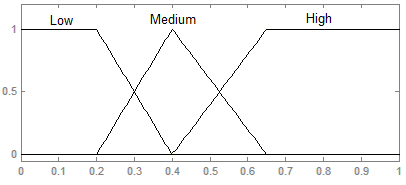}
\caption{Membership functions of the fuzzy sets related to $\rho$ (for dataset A)}
\label{fig:mf-o-ccc}
\end{figure}

Finally, for the fuzzy output $\rho$ we define the following fuzzy sets, whose membership functions are depicted in Figure \ref{fig:mf-o-ccc}:
\begin{itemize}
\item if $\rho(i)\leq 0.4$, then $\rho(i)$ is \emph{Low};
\item if $0.2\leq \rho(i)\leq 0.65$, then $\rho(i)$ is \emph{Medium};
\item if $\rho(i)\geq 0.4$, then $\rho(i)$ is \emph{High};
\end{itemize}

The inference system is based on the following rules, that combine the three inputs $\bar f(i), \bar g(i), \bar h(i)$ in order to produce the fuzzy output $\rho(i)$, for each column $i$ of the matrix of pixels:
\begin{enumerate}
\item if $\bar f(i)$ is not High and $\bar g(i)$ is not High and $\bar h(i)$ is Low, then $\rho(i)$ is Low;
\item if $\bar f(i)$ is Low and $\bar g(i)$ is Low and $\bar h(i)$ is Medium, then $\rho(i)$ is Low;
\item if $\bar f(i)$ is Low and $\bar g(i)$ is High, then $\rho(i)$ is Medium;
\item if $\bar g(i)$ is Medium and $\bar h(i)$ is Medium, then $\rho(i)$ is Medium;
\item if $\bar f(i)$ is High and $\bar g(i)$ is Low, then $\rho(i)$ is Medium;
\item if $\bar f(i)$ is Medium and $\bar g(i)$ is Low and $\bar h(i)$ is Medium, then $\rho(i)$ is Medium;
\item if $\bar f(i)$ is High and $\bar g(i)$ is Medium and $\bar h(i)$ is Low, then $\rho(i)$ is Medium;
\item if $\bar f(i)$ is Medium and $\bar g(i)$ is High, then $\rho(i)$ is High;
\item if $\bar f(i)$ is High and $\bar g(i)$ is High, then $\rho(i)$ is High;
\item if $\bar f(i)$ is High and $\bar g(i)$ is Medium and $\bar h(i)$ is High, then $\rho(i)$ is High.
\end{enumerate}

The touching characters in dataset A are correctly segmented in the $81.1\%$ of the cases. Let us remember that in dataset A we have stored the textual descriptor indicating the index of the proper cut column. We consider a correct segmentation when the routine locates such a column. Let us note that CCC database provides a challenging set of characters for segmentation purposes. The only reference where segmentation of touching characters obtained from CCC database is performed similarly to this section is \cite{Kur} whose authors obtained correct segmentation in the $76.2\%$ of the cases. Let us observe that these results are not directly comparable and are reported just to give a reference of the goodness of our approach, since our dataset A and dataset used in \cite{Kur} are different, even if they are obtained starting from the same CCC database. 
Also, authors in \cite{Kur} took into account percentage of inaccurate segmentation, evaluating when a cutting column is found around the correct position; in this case they report a success percentage of $91.9\%$, without giving further details.
We replicated such an experiment by considering proper, even if inaccurate, segmentation when the cut position identified is distant at most 5 columns from the exact position. In this case, our success percentage is $88.9\%$.
As stated in \cite{lee2012binary}, even though research in handwriting recognition has been an active research area for more than a half century, the maturity of the segmentation techniques is still very low. Our proposed approach focused to improve the segmentation accuracy, achieving comparable, when not better, results.

\begin{figure*}[t!]
    \centering
    \begin{subfigure}[t]{0.3\textwidth}
        \centering
        \includegraphics[width=\textwidth]{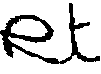}
        \caption{\label{fig:rt}}
    \end{subfigure}%
    ~ 
    \begin{subfigure}[t]{0.3\textwidth}
        \centering
        \includegraphics[width=\textwidth]{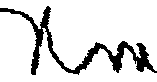}
        \caption{\label{fig:xm}}
    \end{subfigure}%
    ~ 
    \begin{subfigure}[t]{0.3\textwidth}
        \centering
        \includegraphics[width=\textwidth]{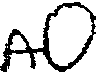}
        \caption{\label{fig:ao}}
    \end{subfigure}
    \caption{\label{fig:3-touch-ccc}Touching characters ``rt'', ``xm'', and ``ao'' extracted from dataset A.}
\end{figure*}

Some numerical results showing the behavior of our method are reported below, where segmentation is performed on touching characters ``eh'', ``'rt'', ``'xm'', and ``ao'' depicted in Figures \ref{fig:eh}, \ref{fig:rt}, \ref{fig:xm}, and \ref{fig:ao}, respectively. 

For the touching characters ``eh'', our fuzzy routine identifies the correct cutting column as the column 52, whereas function $\bar f$ assumes the lowest value in correspondence of the column 56, function $\bar g$ in correspondence of columns 34, 35, 36, and 37, and function $\bar h$ in correspondence of the column 15.

For the touching characters ``rt'', our fuzzy routine identifies the correct cutting column as the column 61, whereas function $\bar f$, $\bar g$, and $\bar h$ identify the cutting position in correspondence of the columns 51, 61, and 70, respectively.

For the touching characters ``xm'', our fuzzy routine identifies the correct cutting column as the column 75, whereas function $\bar f$, $\bar g$, and $\bar h$ identify the cutting position in correspondence of the columns 79, 72, and 137, respectively.

Finally, for the touching characters ``ao'', our fuzzy routine identifies the correct cutting column as the column 43, whereas function $\bar f$ assumes the lowest value in correspondence of columns 48 and 49, function $\bar g$ in correspondence of the column 38, and function $\bar h$ in correspondence of the column 43.

\begin{figure*}[t!]
    \centering
    \begin{subfigure}[t]{0.3\textwidth}
        \centering
        \includegraphics[width=\textwidth]{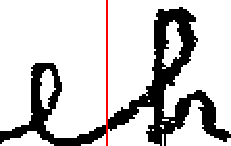}
        \caption{\label{fig:eh-cut}}
    \end{subfigure}%
    ~ 
    \begin{subfigure}[t]{0.3\textwidth}
        \centering
        \includegraphics[width=\textwidth]{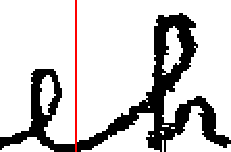}
        \caption{\label{fig:eh-cut-g}}
    \end{subfigure}%
    ~ 
    \begin{subfigure}[t]{0.3\textwidth}
        \centering
        \includegraphics[width=\textwidth]{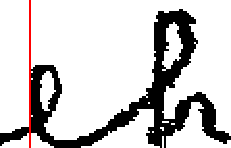}
        \caption{\label{fig:eh-cut-h}}
    \end{subfigure}
    \caption{\label{fig:eh-cuts} Cutting positions located by $\rho$ (\subref{fig:eh-cut}), $\bar g$ (\subref{fig:eh-cut-g}), and $\bar h$ (\subref{fig:eh-cut-h}).}
\end{figure*}

\begin{figure*}[t!]
    \centering
    \begin{subfigure}[t]{0.3\textwidth}
        \centering
        \includegraphics[width=\textwidth]{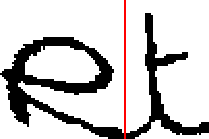}
        \caption{\label{fig:rt-cut}}
    \end{subfigure}%
    ~ 
    \begin{subfigure}[t]{0.3\textwidth}
        \centering
        \includegraphics[width=\textwidth]{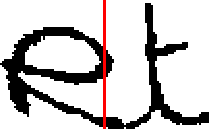}
        \caption{\label{fig:rt-cut-f}}
    \end{subfigure}%
    ~ 
    \begin{subfigure}[t]{0.3\textwidth}
        \centering
        \includegraphics[width=\textwidth]{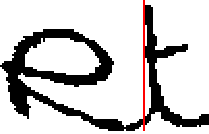}
        \caption{\label{fig:rt-cut-h}}
    \end{subfigure}
    \caption{\label{fig:rt-cuts} Cutting positions located by $\rho$ (\subref{fig:rt-cut}), $\bar f$ (\subref{fig:rt-cut-f}), and $\bar h$ (\subref{fig:rt-cut-h}).}
\end{figure*}

In Figures \ref{fig:eh-cuts} and \ref{fig:rt-cuts}, we show the cutting columns found by $\rho$, $\bar g$, and $\bar h$ related to touching characters ``eh'' and cutting columns found by $\rho$, $\bar f$, $\bar h$ related to touching characters ``rt'', respectively. 

For the sake of readability, we do not report values of $\rho$, $\bar f$, $\bar g$, $\bar h$ for each column since these patterns have usually more than 100 columns.

\section{Conclusion} \label{sec:conc}

A fuzzy approach for segmentation of touching characters has been presented. The proposed method combines three classical features of touching characters usually exploited one at a time. Experiments have been conducted on two very different datasets composed by Latin printed and handwritten characters, respectively. Fuzzy sets, membership functions, and fuzzy rules, which characterize the fuzzy inference scheme, have been properly constructed by means of expert--based choices, heuristic criteria, and the PSO algorithm for each dataset. Numerical results are encouraging and show that the proposed method has an optimal capability of correctly separating touching characters and may be adjusted for very different varieties of characters (not only for the types considered in our experiments). 

Our research activity have shown that even by using alone the functions $f$, $g$, $h$ it is sometime possible to correctly segmenting touching characters. In fact, some experiments (not presented here) have shown that adding other inputs to the fuzzy routine does not provide significant improvements. This is surely a perspective that should be further investigated and motivated. Indeed, looking at perspective advancements, the following issues could be addressed in future works:
\begin{itemize}
\item study of characterization of performance improvements when adding further features as inputs in the fuzzy inference system;
\item experiments on further datasets not involving Latin characters;
\item experiments on segmentation of formulae, taking also into account possibility of segmenting touching characters vertically, horizontally and diagonally;
\item use of fuzzy models different form the Mamdani one (as, e.g., the Sugeno model).
\end{itemize}

\section*{Acknowledgments}
This work has been developed in the framework of an agreement between IRIFOR/UICI (Institute for Research, Education and Rehabilitation/Italian Union for the Blind and Partially Sighted) and Turin University.

This research has been partly supported from the National Thematic Laboratory ``AsTech" of CINI.

Special thanks go to Dr. Tiziana Armano and Prof. Anna Capietto for their support to this work.


\begin{thebibliography}{99}

\bibitem{Alex} V. Alexandrov, \emph{Using critical points in contours for segmentation of touching characters}, Proc. of the 5th Int. Conference on Computer Systems and Technologies, 1--5, New York, 2014.

\bibitem{Ban} V. Bansal, R. M. K. Sinha, \emph{Segmentation of touching and fused Devanagari characters}, Pattern Recognition, Vol. \textbf{35}, 875--893, 2002.

\bibitem{Bayer} T. A. Bayer, U. H. G. Krebel, \emph{Cut classification for segmentation}, IEEE Proc. of 2th International Conference on Document Analysis and Recognition (ICDAR), 565--568, 1993.

\bibitem{bansal2002segmentation} V. Bansal, R. M. K. Sinha, \emph{Segmentation of touching and fused devanagari characters}, Pattern recognition, Vol. \textbf{35}, No. \textbf{4}, 875--893, 2002.

\bibitem{Cama} F. Camastra, M. Spinetti, A. Vinciarelli, \emph{Offline cursive character challange: a new benchmark for machine learning and pattern recognition algorithms}, Proc. of the 18th Int. Conference on Pattern Recognition, Vol. \textbf{2}, 913--916, 2006.

\bibitem{Casey} R. G. Casey, E. Lecolinet, \emph{A survey of methods and strategies in character segmentation}, IEEE Trans. on Pattern Analysis and Machine Intelligence, Vol. \textbf{18}, No. \textbf{7}, 690--706, 1996. 

\bibitem{Fle} L. A. Fletcher, R. Kasturi, \emph{A robust algorithm for text string separation from mixed text-graphics images}, IEEE Trans. on Pattern Analysis and Machine Intelligence, Vol. \textbf{20}, No. \textbf{6}, 910--918, 2002.

\bibitem{Frink} V. Frinken, A. Fischer, R. Mammatha, H. Brunke, \emph{A novel word spotting method based on recurrent neural networks}, IEEE Trans. on Pattern Analysis and Machine Intelligence, Vol. \textbf{34}, No. \textbf{2}, 211--224, 2011.

\bibitem{Utpal} U. Garain, B. B. Chaudhuri, \emph{Segmentation of touching symbols for OCR of printed mathematical expressions: an approach based on multifactorial analysis}, IEEE Proc. of 8th International Conference on Document Analysis and Recognition (ICDAR), Vol. \textbf{1}, 177--181, 2005.

\bibitem{Utpal2} U. Garain, B. B. Chaudhuri, \emph{Segmentation of touching characters in printed Devnagari and Bangla scripts using fuzzy multifactorial analysis}, IEEE Trans. on Systems, Man and Cybernetics, Vol. \textbf{32}, No. \textbf{4}, 449--459, 2002.

\bibitem{He} S. He, M. Wiering, L. Schomaker, \emph{Junction detection in handwritten documents and its application to writer identification}, Pattern Recognition, Vol. \textbf{48}, 4036--4048, 2015.

\bibitem{Heb} J. F. Hebert, M. Parizeau, N. Ghazzali, \emph{Learning to segment cursive words using isolated characters}, Proc. of Conference on Vision Interface, 33--40, 1999.

\bibitem{Jay} M. K. Jasim, A. H. Al--Saleh, A- Aijanaby, \emph{A fuzzy based feature extraction approach for handwritten characters}, International Journal of Computer Science, Vol. \textbf{10}, No. \textbf{4}, 208--215, 2013

\bibitem{Jung} M. C. Jung, Y. C. Shin, S. N. Srihari, \emph{Machine printed character segmentation method using side profiles}, Proceedings of IEEE International Conference onSystems, Man and Cybernetics, New York, 863--867, 1999.

\bibitem{Kahan} S. Kahan, \emph{On the recognition of printed characters of any font and size}, IEEE Transactions on Pattern Analysis and Machine Intelligence, Vol. \textbf{9}, No. \textbf{2}, 274--288, 1987.

\bibitem{PSO} J. Kennedy, R. Eberhart, \emph{Particle swarm optimization}, IEEE Int. Conference on Neural Networks, Perth, Australia, Vol. IV, 1942--1948, 2012.

\bibitem{kim2000recognition} K. K. Kim, J. H. Kim, C. Y. Suen, \emph{Recognition of unconstrained handwritten numeral strings by composite  segmentation method} In Pattern Recognition, 2000. Proceedings. 15th International Conference on, Vol. \textbf{2}, 594--597. IEEE, 2000.

\bibitem{Kur} F. Kurniawan, M. S. M. Rahim, D. Daman, A. Rehman, D. Mohamad, S. M. Shamsuddin, \emph{Region--based touched character segmentation in handwritten words}, International Journal of Innovative Computing, Information and Control, Vol. \textbf{7}, No. \textbf{6}, 3107--3120, 2011.

\bibitem{Kumar} M. Kumar, M. K. Jindal, R. K. Sharma, \emph{Segmentation of isolatedtouching characters in offline handwritten Grumukhi script recognition}, Int. J. of Information Technology and Computer Science, Vol. \textbf{2}, 58--63, 2014.

\bibitem{LeeVerma} H. Lee, B. Verma, \emph{Binary segmentation algorithm for English cursive handwriting recognition}, Pattern Recognition, Vol. \textbf{45}, 1306--1317, 2012.

\bibitem{Liang} J. Liang, I. T. Phillips, R. M. Haralick, \emph{An optimization methodology for document structure extraction on Latin character documents}, IEEE Trans. on Pattern Analysis and Machine Intelligence, Vol. \textbf{23}, No. \textbf{7}, 719--734, 2002.

\bibitem{liang1994segmentation} S. Liang, M. Shridhar, M. Ahmadi, \emph{Segmentation of touching characters in printed document recognition}, Pattern Recognition, Vol. \textbf{27}, No. \textbf{6}, 825--840, 1994.

\bibitem{Lou} G. Louloudis, B. Gatos, I. Pratikakis, C. Halatsis, \emph{Text line and word segmentation of handwritten documents}, Pattern Recognition, Vol. \textbf{42}, 3169--3183, 2009.

\bibitem{Lu} Y. Lu, \emph{Machine printed character segmentation -- an overview}, Pattern Recognition, Vol. \textbf{28}, No. \textbf{1}, 67--80, 1995.

\bibitem{Lu4} Y. Lu, \emph{On the segmentation of touching characters}, IEEE Proc. of 2th International Conference on Document Analysis and Recognition (ICDAR), 440--443, 1993.

\bibitem{Lu3} Z. Lu, Z. Chi, W. C. Siu, P. Shi, \emph{A background--thinning--based approach for separating and recognizing connected handwritten digit strings}, Pattern Recognition, Vol. \textbf{32}, 921--933, 1999.

\bibitem{Lu2} Y. Lu, M. Shridhar, \emph{Character segmentation in handwritten words -- an overview}, Pattern Recognition, Vol. \textbf{29}, No. \textbf{1}, 77--96, 1996.

\bibitem{Mam} E. H. Mamdani, S. Assilian, \emph{An experiment in linguistic synthesis with a fuzzy logic controller}, International Journal of Man--Machine Studies, Vol. \textbf{7}, No. \textbf{1}, 1--13, 1975.

\bibitem{Manma} R. Manmatha, J. L. Rothfeder, \emph{A scale space approach for automatically segmenting words from historical handwritten documents} IEEE Trans. on Pattern Analysis and Machine Intelligence, Vol. \textbf{27}, No. \textbf{8}, 1212--1225, 2005. 

\bibitem{Mon} T. Mondal, N. Ragot, J. Y. Ramel, U. Pal, \emph{Flexible sequence matching technique: an effective learning--free approach for word spotting}, Pattern Recognition, Vol. \textbf{60}, 596--612, 2016.

\bibitem{Nac} R. Nachar, E. Inaty, P. J. Bonnin, Y. Alayli, \emph{Breaking down Captcha using edge corners and fuzzy logic segmentation/recognition technique}, Security and Communication Networks, Vol. \textbf{8}, 3995--4012, 2015.

\bibitem{Naz} S. Naz, H. Majeed, H. Irshad, \emph{Image segmentation using fuzzy clustering: a survey}, Proc. of 6th International Conference on Emerging Technologies (ICET), 181--186, 2010.

\bibitem{Infty} A. Nomura, K. Michishita, S. Uchida, M. Suzuki, \emph{Detection and segmentation of touching characters in mathematical expressions}, IEEE Proc. of 7th International Conference on Document Analysis and Recognition (ICDAR), Vol. \textbf{1}, 126--130, 2003.

\bibitem{Otsu} N. Otsu, \emph{A treshold selection method from gray--level histograms}, IEEE Trans. Sys., Man., Cyber., Vol. \textbf{9}, No. \textbf{1}, 62--66, 1979.

\bibitem{Pal} U. Pal, A. Belad, C. Choisy, \emph{Touching numeral segmentation using water reservoir concept}, Pattern Recognition Letters, Vol. \textbf{24}, 261--272, 2003.

\bibitem{Pra} S. Pravesjit, A. Thammano, \emph{Touching character segmentation method of archaic Lanna script}, Chapter E--business and Telecommunication, Vol. \textbf{314}, Series Communications in Computer and Information Science, 400--408, 2012.

\bibitem{Reh} A. Rehman, F. Kurniawan, D. Mohamad, \emph{Off--line cursive handwriting segmentation: a heuristic rule--based approach}, Journal of Institute of Mathematics and Computer Science, Vol. \textbf{19}, No. \textbf{2}, 135--140, 2008.

\bibitem{Roy} P. P. Roy, U. Pal, J. Llados, \emph{Proccedings of the IEEE SiRecognition of multi--oriented touching characters in graphical documents}, Proceedings of the IEEE Sixth Indian Conference on Computer Vision, Graphics and Image Processing, 297--304, 2008.

\bibitem{roy2012multi} P. P. Roy, U. Pal, J. Llad{\'o}s, M. Delalandre, \emph{Multi-oriented touching text character segmentation in graphical documents using dynamic programming}, Pattern Recognition, Vol. \textbf{45}, No. \textbf{5}, 1972--1983, 2012.

\bibitem{Roy2} P. P. Roy, U. Pal, J. Llados, M. Delandre, \emph{Multi--oriented touching character segmentation in graphical documents using dynamic programming}, Pattern Recognition, Vol. \textbf{45}, 1972--1983, 2012.

\bibitem{Saba} T. Saba, G. Sulong, A. Rehman, \emph{A survey on methods and strategies on touched characters segmentation}, International Journal of Research and Reviews in Computer Science, Vol. \textbf{2}, No. \textbf{1}, 103--114, 2010.

\bibitem{Saba2} T. Saba, G. Sulong, A. Rehman, \emph{Non--linear segmentation of touched Roman characters based on genetic algorithm}, Int. J. on Computer Science and Engineering, Vol. \textbf{2}, No. \textbf{6}, 2167--2172, 2010. 

\bibitem{sadri2016novel} J. Sadri, M. R. Yeganehzad, J. Saghi, \emph{A novel comprehensive database for offline persian handwriting recognition}, Pattern Recognition,  Vol. \textbf{60}, 378--393, 2016.

\bibitem{Sarkar} R. Sarkar, B. Sen, N. Das, S. Basu, \emph{Handwritten Devanagari script segmentation: a non--linear fuzzy approach}, Proc. of IEEE Conference on AI Tools and Engineering (ICAITE), 2008.

\bibitem{Sha} N. Sharma, P. Shvakumara, U. Pal, M. Blumenstein, C. L. Tan, \emph{A new method for character segmentation from multi--oriented video words}, IEEE Proc. of 12th International Conference on Document Analysis and Recognition (ICDAR), Vol. \textbf{1}, 413--417, 2013.

\bibitem{Shi} Z. Shi, V. Govindaraju, \emph{Line separation for complex document images using fuzzy runlength}, IEEE Proc. of the 1st International Workshop on Document Image Analysis for Libraries, 306--312, 2004.

\bibitem{Sta} N. Stamatopoulos, B. Gatos, S. J. Perantonis, \emph{A method for combining complementary techniques for document image segmentation}, Pattern Recognition, Vol. \textbf{42}, 3158--3168, 2009.

\bibitem{Tobias} O. J. Tobias, R. Seara, \emph{Image segmentation by histogram thresholding using fuzzy sets}, IEEE Trans. on Image Processing, Vol. \textbf{11}, No. \textbf{12}, 2002.

\bibitem{lee2012binary} H. Lee, B. Verma, \emph{Binary segmentation algorithm for english cursive handwriting  recognition}, Pattern Recognition, Vol. \textbf{45}, No. \textbf{4}, 1306--1317, 2012.

\bibitem{Wei} X. Wei, S. Ma, Y. Jin, \emph{Segmentation of connected Chinese characters based on genetic algorithm}, IEEE Proc. of 8th International Conference on Document Analysis and Recognition (ICDAR), Vol. \textbf{2}, 645--649, 2005.

\bibitem{Weinman} J. J. Weinman, E. L. Miller, A. R. Hanson, \emph{Text recognition using similarity and lexicon with sparse belief propagation}, IEEE Pattern Analysis and Machine  Intelligence, Vol. \textbf{31}, 1733--1746, 2009.

\bibitem{zhao2003two} S. Zhao, Z. Chi, P. Shi, H. Yan, \emph{Two-stage segmentation of unconstrained handwritten chinese characters}, Pattern Recognition, Vol. \textbf{36}, No. \textbf{1}, 145--156, 2003.

\bibitem{zhou2002verification} J. Zhou, A. Krzyzak, C. Y. Suen, \emph{Verification--a method of enhancing the recognizers of isolated and touching handwritten numerals}, Pattern Recognition, Vol. \textbf{35}, No. \textbf{5}, 1179--1189, 2002.

\end{thebibliography}

\end{document}